\documentclass{article}
\usepackage{spconf,amsmath,graphicx,hyperref}
\usepackage{multirow}
\usepackage{threeparttable}
\usepackage{booktabs}  
\usepackage{amssymb} 
\hypersetup{hidelinks}

\begin{document}
\ninept

\title{EASE: Entropy-Adaptive Distribution Shaping  \\ for Evading AI-Generated Text Detectors}

\name{
Jicheng Zhou, 
Kahim Wong,
Jialong Wang,
and Jiantao Zhou\sthanks{Corresponding author (jtzhou@um.edu.mo).}
}

\address{
Department of Artificial Intelligence, Faculty of Information Science and Computing\\
State Key Laboratory of Internet of Things for Smart City \\
University of Macau, Macau, China
}

\maketitle

\begingroup
\renewcommand{\thefootnote}{}
\footnotetext{Code is available at: \url{https://github.com/RupertZJC/EASE}}
\endgroup

\begin{abstract}
AI-generated text (AIGT) detection can be sensitive to the decoding choices of the source large language model (LLM).
We observe that perturbing next-token logits or adjusting sampling temperature can reduce detection performance, providing a clear signal of detector vulnerability to decoding-time distribution changes.
Building on this observation, we propose \textbf{EASE} (\textbf{E}ntropy-\textbf{A}daptive Distribution \textbf{S}haping for \textbf{E}vasion), a training-free and detector-agnostic framework for evading AIGT detectors.
EASE computes predictive entropy directly from the source LLM's next-token distribution and uses it to adapt both logit perturbation and sampling temperature, without detector feedback or model fine-tuning.
Experiments across three source LLMs and multiple detectors demonstrate consistent reductions in detection performance, with negligible degradation in text quality and negligible inference overhead.
\end{abstract}

\begin{keywords}
  AIGT Detection Evasion, Adversarial Attack, LLM, Information Forensics
\end{keywords} 

\section{Introduction}
Large language models (LLMs) generate text whose statistical characteristics
can differ from those of human-written text.
AI-generated text (AIGT) detection is an important task in information forensics  seeking to distinguish machine-generated text from human-written text.
Related information-forensics tasks include AI-generated image detection~\cite{ntire2026,featdistill2026}, image provenance analysis~\cite{editprint2026}, and visual-text forgery detection~\cite{seed2026,rise2026}.
Existing approaches include training-free statistical
detectors~\cite{lavergne2008detecting,gehrmann2019gltr,mitchell2023detectgpt,
bao2024fast,yang2024dna,su2023detectllm,wong2026k},
supervised classifiers~\cite{solaiman2019release,liu2019roberta,
li2305mage,hu2023radar},
and watermark-based methods~\cite{kirchenbauer2023watermark,wong2025end,wong2025fontguard}.
Recent work has also explored multi-LLM statistical
detection~\cite{mao2025mlsdet} and in-the-wild AIGT
benchmarking~\cite{zhang2026detectwild}.
The statistical characteristics of AIGT depend in part on the source LLM's decoding choices and output distributions~\cite{li2026llm}.

\begin{figure}[t]
\centering
\includegraphics[width=\linewidth]{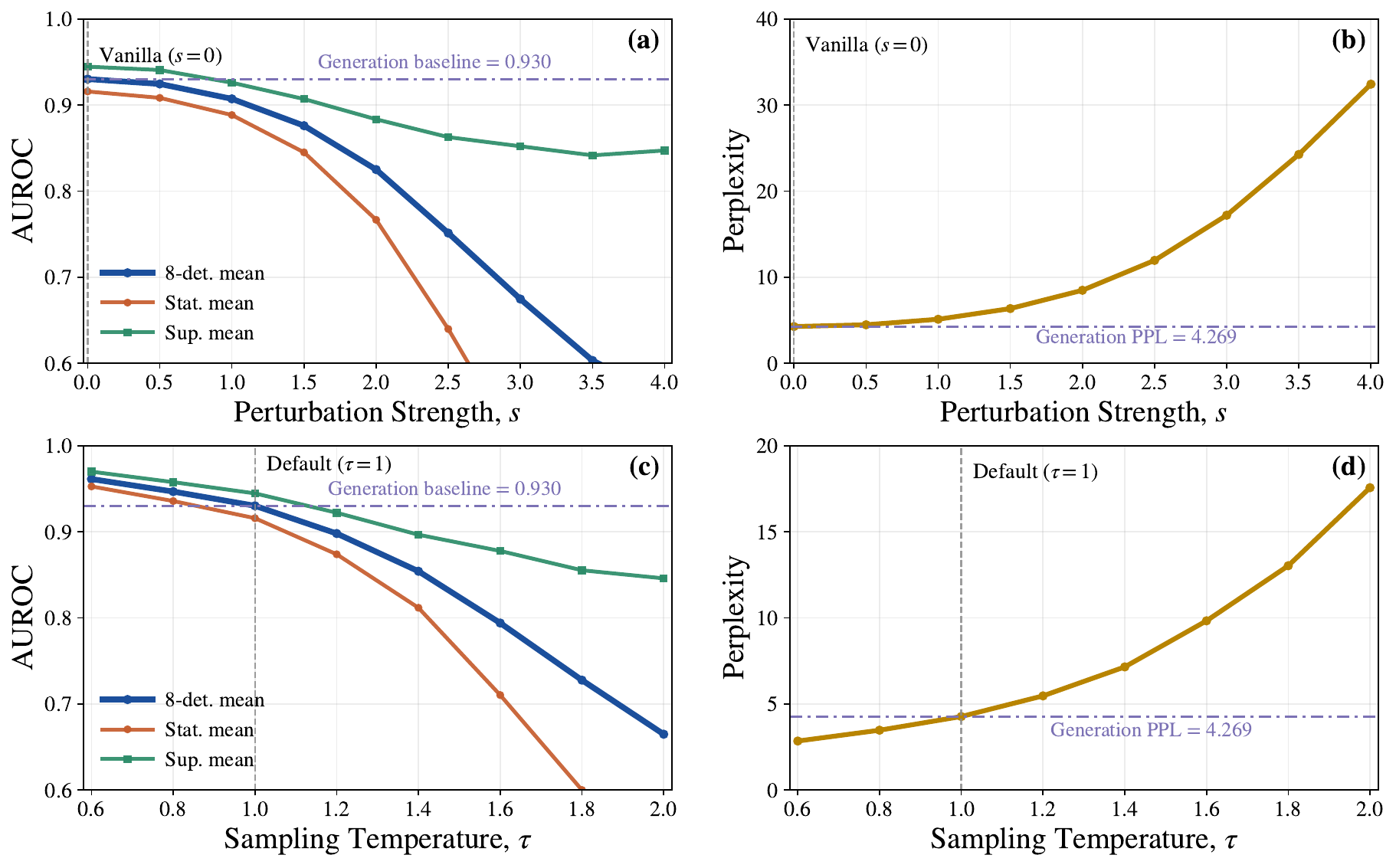}
\vspace{-15pt}
\caption{
\textbf{Effects of decoding changes on AIGT detection and text quality.}
AUROC and perplexity under (a,b) top-$k$ logit perturbation with
strength $s$ at $\tau=1$ and (c,d) varying sampling temperature
without logit perturbation.
AUROC curves show means over all eight detectors and within two groups:
statistical (Likelihood~\cite{lavergne2008detecting},
Entropy and LogRank~\cite{gehrmann2019gltr},
and LRR~\cite{su2023detectllm})
and supervised (RoBERTa-Base/Large~\cite{liu2019roberta},
MAGE~\cite{li2305mage}, and RADAR~\cite{hu2023radar}).
Dash-dotted lines mark vanilla baselines.
}
\vspace{-10pt}
\label{fig:motivation}
\end{figure}

As AIGT detection advances, recent work has explored methods for evading
AIGT detectors while preserving text quality.
Existing methods mainly follow two directions: text-level modification
and direct modification of the source LLM's generation process.
At the text level, paraphrasing remains a dominant
strategy~\cite{sadasivan2023can,cheng2026adversarial}.
Other methods include in-context optimization~\cite{lu2023large}
and space infiltration~\cite{cai2023evade}.
Methods that directly modify the generation process include preference
optimization~\cite{rafailov2023direct,nicks2024language}
and proxy-guided decoding~\cite{wang2025humanizing}.
Depending on their design, these methods may require detector feedback,
auxiliary models, or model fine-tuning.
These requirements introduce additional computation and may limit their
applicability across different detectors or source models.
This raises the question: \emph{can AIGT detectors be evaded directly
during generation, without detector feedback or model fine-tuning?}

\begin{figure*}[!t]
\centering
\includegraphics[width=\textwidth]{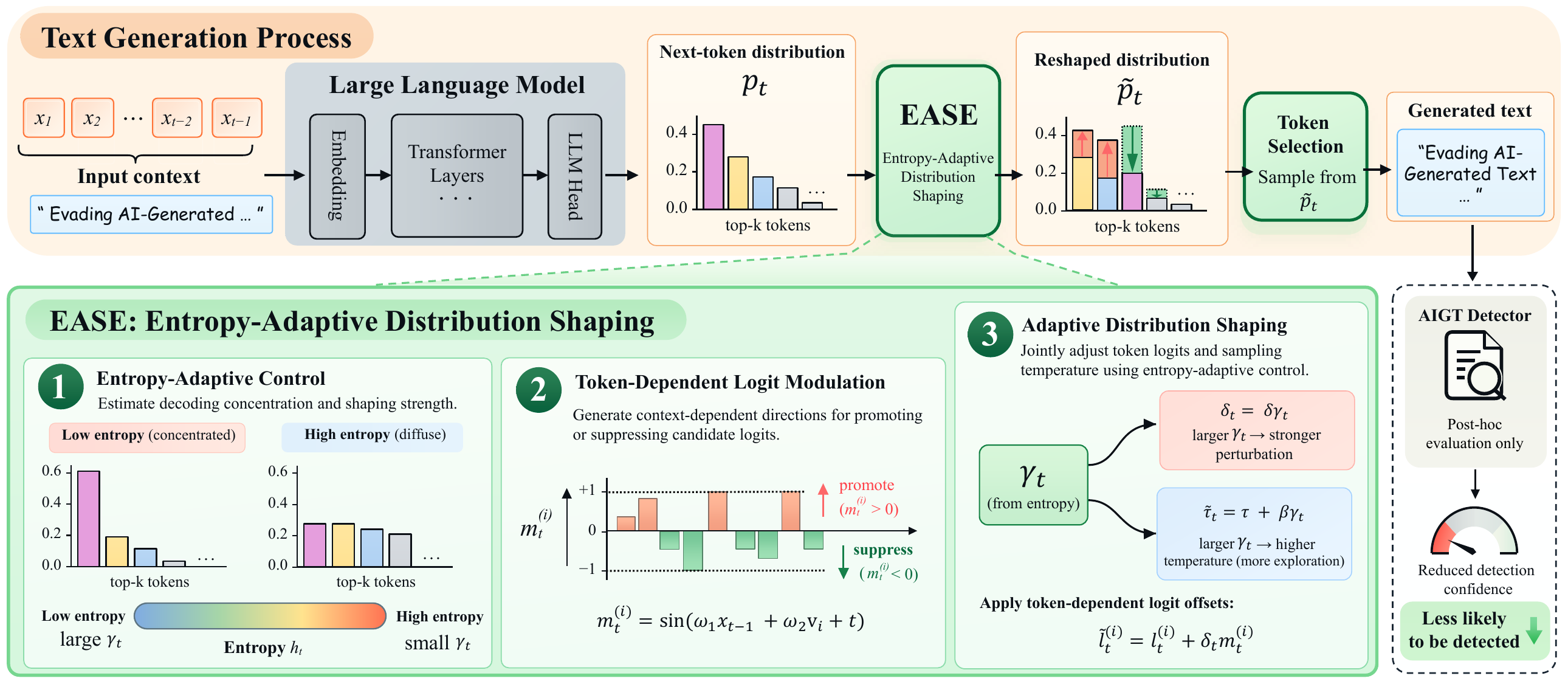}
\vspace{-10pt}
\caption{
\textbf{Overview of EASE.}
EASE uses top-$k$ predictive entropy to adapt logit perturbation
strength and sampling temperature, together with token-dependent
logit modulation to reshape the next-token distribution before sampling.
}
\vspace{-10pt}
\label{fig:overview}
\end{figure*}

To investigate this question, we study how changes in the source LLM's
decoding distribution influence downstream AIGT detection.
As shown in Fig.~\ref{fig:motivation}, perturbing next-token logits or
adjusting sampling temperature can reduce detector AUROC, providing
a clear signal of detector vulnerability to decoding-time distribution changes.
The accompanying increase in perplexity also indicates potential
text quality degradation.
Across decoding steps, probability may be concentrated on a few
candidate tokens or distributed more evenly among them.
Fixed logit perturbation strengths and sampling temperatures do not
account for this variation.
To account for these differences, we consider adapting both operations
to the concentration of the current candidate distribution.
Predictive entropy offers one way to quantify this concentration,
with lower entropy indicating a more concentrated distribution.

Building on this observation, we introduce \textbf{EASE}
(\textbf{E}ntropy-\textbf{A}daptive Distribution \textbf{S}haping for
\textbf{E}vasion), a training-free and detector-agnostic framework for
evading AIGT detectors.
As illustrated in Fig.~\ref{fig:overview}, EASE computes predictive entropy
over the source LLM's top-$k$ candidate tokens and uses it to adapt
logit perturbation and sampling temperature, without detector feedback
or model fine-tuning.
At lower entropy, EASE applies larger logit perturbations and higher
sampling temperatures to encourage alternative token selections.
As the candidate distribution approaches uniformity, these adjustments
diminish.
On Qwen3-8B~\cite{yang2025qwen3}, EASE reduces the average
TPR@1\%FPR across five detectors from 0.655 to 0.339, with a measured
generation latency overhead of 0.643\%.

Our key contributions are as follows:
\begin{itemize}
\item We reveal that both logit perturbation and temperature scaling can reduce
AIGT detector performance, providing a clear signal of detector
vulnerability.

\item We propose EASE, a training-free and detector-agnostic
framework for evading AIGT detectors.
EASE uses predictive entropy to adapt logit perturbation and
sampling temperature without detector feedback or model fine-tuning.

\item Experiments across three source LLMs and multiple detectors
demonstrate consistent reductions in detection performance with
negligible inference overhead and quantify the associated changes
in text quality.
\end{itemize}

\section{Decoding-Time Distribution Changes Affect Downstream Detection}
\label{sec:motivation}
We first formulate the decoding process of a standard autoregressive LLM.
Let $\mathcal{V}$ denote the vocabulary. At generation step $t$, given the
preceding context $\mathbf{x}_{<t}=[x_1,\dots,x_{t-1}]$, the model outputs
a logits vector $\mathbf{l}_t \in \mathbb{R}^{|\mathcal{V}|}$.
The next token is sampled from
\begin{equation}
p(x_t=v_i \mid \mathbf{x}_{<t})
= p_t^{(i)}
= \frac{\exp(l_t^{(i)}/\tau)}
{\sum_{j=1}^{|\mathcal{V}|}\exp(l_t^{(j)}/\tau)},
\end{equation}
where $v_i\in\mathcal{V}$ and $\tau>0$ controls the concentration of the
next-token distribution. 

We conduct a small-scale observation study using the setup described
in Section~\ref{sec:exp_setup}, with a different random seed.
First, we add Gaussian noise to the top-$k$ logits:
$\hat{l}_t^{(i)}=l_t^{(i)}+s\epsilon_t^{(i)}$ for $i\in\mathcal{K}_t$,
where $\mathcal{K}_t$ indexes the top-$k$ unperturbed logits
and sampling is restricted to this set,
$\epsilon_t^{(i)}\overset{\mathrm{i.i.d.}}{\sim}\mathcal{N}(0,1)$,
and $s\geq0$ controls the perturbation strength. 
We vary $s$ at fixed $\tau=1$.
As shown in Fig.~\ref{fig:motivation}(a,b), stronger perturbations
generally reduce detector AUROC but also increase generation perplexity.
Second, we vary $\tau$ without logit perturbation.
Fig.~\ref{fig:motivation}(c,d) shows a similar pattern:
higher temperatures can improve evasion while increasing perplexity.

These observations provide a clear signal of detector vulnerability
to decoding-time distribution changes.
However, stronger logit perturbations or higher sampling temperatures
can also increase perplexity.
Fixed settings do not account for variations in the concentration
of the candidate distribution across decoding steps.
We therefore consider adapting logit perturbation and sampling
temperature to the current distribution.
Predictive entropy offers one way to quantify its concentration,
providing a basis for the entropy-adaptive design of EASE
in the following section.

\section{Entropy-Adaptive Distribution Shaping}
\label{sec:method}

We introduce EASE, a framework for evading AIGT detectors during generation.
As illustrated in Fig.~\ref{fig:overview}, EASE uses predictive entropy
to adapt logit perturbation strength and sampling temperature.
A token-dependent modulation determines the sign and relative magnitude
of the logit perturbation for each candidate.
These operations reshape the source LLM's next-token distribution
before sampling, without detector feedback or model fine-tuning.

\begin{table*}[!t]
\centering
\setlength{\tabcolsep}{5.5pt}
\caption{
Comparison with existing evasion methods on Qwen3-8B.
Each detector entry reports AUROC / TPR@1\%FPR.
Lower values indicate stronger evasion; lower PPL indicates a higher likelihood under the source LLM.
}
\vspace{8pt}
\label{tab:main_competitors}

\begin{tabular}{@{}lcccccc@{}}
\toprule
& \multicolumn{1}{c}{\textbf{Text Quality}}
& \multicolumn{5}{c}{\textbf{Detection Performance}} \\
\cmidrule(lr){2-2}
\cmidrule(lr){3-7}

\textbf{Method}
& \textbf{PPL}
& \textbf{RoBERTa-Large}
& \textbf{RoBERTa-Base}
& \textbf{Fast-DetectGPT}
& \textbf{MAGE}
& \textbf{RADAR} \\
\midrule

No Attack
& 4.243
& 0.921 / 0.535
& 0.930 / 0.580
& 0.984 / 0.816
& 0.981 / 0.685
& 0.943 / 0.660 \\

Simple Paraphrase
& 12.120
& 0.950 / 0.615
& 0.961 / 0.695
& 0.896 / 0.452
& 0.912 / 0.400
& 0.884 / 0.568 \\

Rec. Para. 2
& 9.867
& 0.942 / 0.602
& 0.959 / 0.643
& 0.858 / 0.391
& 0.897 / 0.353
& 0.870 / 0.536 \\

AdvPara (RoBERTa-L) ~\cite{cheng2026adversarial}
& 15.088
& 0.946 / 0.595
& 0.961 / 0.664
& 0.897 / 0.455
& 0.915 / 0.399
& 0.882 / 0.567 \\

AdvPara (RoBERTa-B) ~\cite{cheng2026adversarial}
& 15.196
& 0.939 / 0.583
& 0.954 / 0.584
& 0.893 / 0.435
& 0.912 / 0.398
& 0.881 / 0.567 \\

\midrule

\textbf{EASE-Rewrite (Ours)}
& 13.370
& 0.918 / 0.507
& 0.936 / 0.566
& \textbf{0.845} / 0.296
& \textbf{0.884 / 0.303}
& \textbf{0.864} / 0.536 \\

\textbf{EASE-Plugin (Ours)}
& \textbf{7.752}
& \textbf{0.828 / 0.250}
& \textbf{0.839 / 0.260}
& 0.923 / \textbf{0.295}
& 0.911 / 0.445
& 0.895 / \textbf{0.445} \\

\bottomrule
\vspace{-15pt}
\end{tabular}
\end{table*}

\subsection{Entropy-Adaptive Control}
\label{sec:entropy_control}

At generation step $t$, we compute predictive entropy from the
unmodified next-token distribution over the top-$k$ candidate set
$\mathcal{K}_t$ defined in Section~\ref{sec:motivation}.
We renormalize the next-token probabilities over this set as
$q_t^{(i)} = p_t^{(i)} /
\sum_{j\in\mathcal{K}_t} p_t^{(j)}$,
$i\in\mathcal{K}_t$, and compute
\begin{equation}
h_t = -\sum_{i\in\mathcal{K}_t}
q_t^{(i)} \log q_t^{(i)}.
\end{equation}
Using the maximum entropy $h_{\max}=\log k$, we define
\begin{equation}
\gamma_t =
\max\left(0,\,1-\frac{h_t}{h_{\max}}\right).
\end{equation}
The factor $\gamma_t \in [0,1]$ controls logit perturbation strength
and sampling temperature.
Low-entropy distributions concentrate probability on a few candidates.
In these states, a larger $\gamma_t$ produces stronger logit perturbations
and higher sampling temperatures to encourage alternative token selections.
As the candidate distribution approaches uniformity, $\gamma_t$
approaches zero, and both adjustments diminish.
This design adapts logit perturbation and sampling temperature
to the current distribution rather than keeping them fixed
throughout generation.

\subsection{Token-Dependent Logit Modulation}
\label{sec:directional_modulation}
Inspired by decoding-time distribution manipulation~\cite{pang2025language},
we assign token-dependent modulation to individual candidates.
A sinusoidal function offers one feasible construction:
\begin{equation}
m_t^{(i)}
=
\sin\left(w_1 x_{t-1} + w_2 v_i + t\right),
\qquad i\in\mathcal{K}_t,
\end{equation}
where $x_{t-1}$ and $v_i$ denote the preceding and candidate
token IDs, respectively, and $w_1,w_2$ are fixed coefficients.
Each modulation coefficient $m_t^{(i)} \in [-1,1]$ determines the
sign and relative magnitude of a candidate's logit offset,
while $\gamma_t$ controls the overall logit perturbation strength. 
This deterministic construction produces context- and token-dependent
offsets without learning additional parameters or querying a detector.

\subsection{Logits Distribution Shaping}
\label{sec:logits_shaping}

EASE combines the entropy-based control factor with the
token-dependent mask to reshape candidate logits.
For each $i\in\mathcal{K}_t$, we apply
$\tilde{l}_t^{(i)}=l_t^{(i)}+\delta\gamma_t m_t^{(i)}$,
where $\delta$ controls the base perturbation magnitude.
The same control factor adjusts the sampling temperature as
$\tilde{\tau}_t=\tau+\beta\gamma_t$,
where $\tau>0$ is the base temperature and $\beta\geq0$
sets the maximum temperature increase.
The next token is sampled from
\begin{equation}
\tilde{p}_t^{(i)}
=
\frac{\exp(\tilde{l}_t^{(i)}/\tilde{\tau}_t)}
{\sum_{j\in\mathcal{K}_t}
\exp(\tilde{l}_t^{(j)}/\tilde{\tau}_t)},
\qquad i\in\mathcal{K}_t,
\end{equation}
with zero probability assigned to tokens outside
$\mathcal{K}_t$.
The two operations play distinct roles: additive modulation
can change the ranking of candidate logits, whereas
temperature adjustment changes distribution concentration
without altering the ranking of the shaped logits.
Both are governed by $\gamma_t$, coupling token-dependent
redistribution with uncertainty-dependent temperature control.

\section{Experiments}
\label{sec:experiments}

\subsection{Experimental Setup}
\label{sec:exp_setup}

\textbf{Attack setting and data.}
We consider a model deployer who controls the source LLM's
decoding process and aims to make its generated text less
detectable by downstream AIGT detectors.
The deployer can modify next-token logits and sampling
temperature, without querying detectors or fine-tuning the model.
We sample 2,000 human-written sequences from
WikiText-103~\cite{merity2016pointer}, using the first 30 tokens
as prompts to generate continuations of up to 200 tokens.
Vanilla and EASE share all other generation settings.

\textbf{Models and baselines.}
The main comparison and ablation use Qwen3-8B~\cite{yang2025qwen3};
cross-detector and quality evaluations additionally use
Llama3-8B-Instruct~\cite{grattafiori2024llama} and
Ministral-3-8B-Instruct~\cite{ministral3}.
Baselines include simple and recursive paraphrasing following
Cheng et al.~\cite{cheng2026adversarial}, building on prior
paraphrasing attacks~\cite{sadasivan2023can,krishna2023paraphrasing},
as well as the AdvPara variants~\cite{cheng2026adversarial}.

\textbf{Detectors and metrics.}
We evaluate RoBERTa-Base/Large~\cite{liu2019roberta},
Fast-DetectGPT~\cite{bao2024fast}, MAGE~\cite{li2305mage},
RADAR~\cite{hu2023radar}, and eight additional detectors in
Table~\ref{tab:cross_detector}.
Detection is measured by AUROC and TPR@1\%FPR, with lower
values indicating stronger evasion under fixed score orientations.
We compute PPL using the corresponding source LLM for generation
quality, and report BLEU-1/2~\cite{papineni2002bleu} and
ROUGE-1/2~\cite{lin2004rouge} for reference overlap.

\textbf{Configurations.}
EASE-Plugin applies logit perturbation and sampling temperature
adjustment during initial generation.
EASE-Rewrite applies the same operations during a subsequent
rewriting stage, matching the generate-then-rewrite workflow
of paraphrasing baselines.
Table~\ref{tab:main_competitors} evaluates both configurations;
all subsequent experiments use EASE-Plugin.
We use top-$k$ sampling with $k=20$ and base temperature $\tau=1$,
and set $\delta=2$ and $\beta=0.5$ for EASE.

\begin{table*}[!t]
\centering
\setlength{\tabcolsep}{6.5pt}
\caption{
Cross-detector evasion across three source LLMs.
Each entry reports AUROC / TPR@1\%FPR; lower is better.
}
\vspace{8pt}
\label{tab:cross_detector}

\begin{tabular}{lcccccc}
\toprule
& \multicolumn{2}{c}{\textbf{Qwen3-8B}}
& \multicolumn{2}{c}{\textbf{Llama3-8B}}
& \multicolumn{2}{c}{\textbf{Ministral-3-8B}} \\
\cmidrule(lr){2-3}
\cmidrule(lr){4-5}
\cmidrule(lr){6-7}

\textbf{Detector}
& \textbf{Vanilla} & \textbf{EASE}
& \textbf{Vanilla} & \textbf{EASE}
& \textbf{Vanilla} & \textbf{EASE} \\
\midrule

Likelihood~\cite{lavergne2008detecting}
& 0.966 / 0.485 & 0.803 / 0.080
& 0.973 / 0.640 & 0.737 / 0.045
& 0.847 / 0.080  & 0.419 / 0.000 \\

Entropy~\cite{gehrmann2019gltr}
& 0.761 / 0.190 & 0.576 / 0.040
& 0.788 / 0.145 & 0.602 / 0.055
& 0.638 / 0.070  & 0.436 / 0.000 \\

LogRank~\cite{gehrmann2019gltr}
& 0.973 / 0.535 & 0.856 / 0.100
& 0.980 / 0.705 & 0.810 / 0.080
& 0.882 / 0.085 & 0.588 / 0.000 \\

LRR~\cite{su2023detectllm}
& 0.963 / 0.745 & 0.919 / 0.530
& 0.959 / 0.805 & 0.917 / 0.515
& 0.917 / 0.485 & 0.896 / 0.365 \\

NPR~\cite{su2023detectllm}
& 0.645 / 0.025 & 0.498 / 0.005
& 0.788 / 0.065 & 0.523 / 0.010
& 0.713 / 0.010 & 0.411 / 0.000 \\

DNA-GPT~\cite{yang2024dna}
& 0.825 / 0.440 & 0.686 / 0.175
& 0.801 / 0.240 & 0.583 / 0.085
& 0.534 / 0.030 & 0.466 / 0.005 \\

DetectGPT~\cite{mitchell2023detectgpt}
& 0.402 / 0.005 & 0.364 / 0.005
& 0.501 / 0.010 & 0.368 / 0.005
& 0.448 / 0.005 & 0.326 / 0.000 \\

Binoculars~\cite{hans2024spotting}
& 1.000 / 0.975 & 0.977 / 0.690
& 0.997 / 0.955 & 0.873 / 0.375
& 0.896 / 0.365 & 0.537 / 0.000 \\

\midrule
\textbf{Mean (8 detectors)}
& 0.817 / 0.425 & \textbf{0.709 / 0.203}
& 0.848 / 0.446 & \textbf{0.676 / 0.146}
& 0.734 / 0.141 & \textbf{0.510 / 0.046} \\

\bottomrule
\vspace{-15pt}
\end{tabular}
\end{table*}

\subsection{Comparison with Existing Evasion Methods}
\label{sec:competitor_results}
Table~\ref{tab:main_competitors} compares EASE with existing
paraphrasing-based evasion methods on Qwen3-8B.
EASE-Plugin reduces both detection metrics across all five detectors
relative to vanilla generation.
The average AUROC decreases from 0.952 to 0.879, while the
average TPR@1\%FPR decreases from 0.655 to 0.339.
It achieves the lowest AUROC and TPR@1\%FPR on both RoBERTa
detectors.
On RoBERTa-Large, for example, TPR@1\%FPR decreases from
0.535 to 0.250, whereas the evaluated paraphrasing baselines
yield values between 0.583 and 0.615.
EASE-Plugin also achieves the lowest PPL among the evaluated
evasion methods at 7.752, compared with 9.867-15.196 for
the paraphrasing baselines.
These results support its effectiveness during initial generation
without requiring post-generation rewriting.

EASE-Rewrite follows the generate-then-rewrite workflow of the
paraphrasing baselines and achieves the lowest AUROC on
Fast-DetectGPT, MAGE, and RADAR.
On MAGE, it also achieves the lowest TPR@1\%FPR at 0.303,
compared with 0.445 for EASE-Plugin.
The AUROC advantage of rewriting does not consistently
translate into stronger evasion at a low false-positive rate.
The relative benefits of the two EASE settings depend
on the detector and evaluation metric.

\subsection{Cross-Detector Evasion and Text Quality}
\label{sec:generalization_quality}

\begin{table}[t]
\centering
\setlength{\tabcolsep}{2.0pt}
\caption{
Text quality under vanilla decoding and EASE across three source LLMs.
}
\vspace{8pt}
\label{tab:text_quality}

\begin{tabular}{lcccccc}
\toprule
& \multicolumn{2}{c}{\textbf{Qwen3-8B}}
& \multicolumn{2}{c}{\textbf{Llama3-8B}}
& \multicolumn{2}{c}{\textbf{Ministral-3-8B}} \\
\cmidrule(lr){2-3}
\cmidrule(lr){4-5}
\cmidrule(lr){6-7}

\textbf{Metric}
& \textbf{Vanilla} & \textbf{EASE}
& \textbf{Vanilla} & \textbf{EASE}
& \textbf{Vanilla} & \textbf{EASE} \\
\midrule

BLEU-1 $\uparrow$
& 0.224 & 0.220
& 0.234 & 0.227
& 0.321 & 0.316 \\

BLEU-2 $\uparrow$
& 0.120 & 0.112
& 0.128 & 0.115
& 0.131 & 0.118 \\

ROUGE-1 $\uparrow$
& 0.364 & 0.368
& 0.375 & 0.363
& 0.337 & 0.338 \\

ROUGE-2 $\uparrow$
& 0.137 & 0.124
& 0.147 & 0.120
& 0.059 & 0.049 \\

PPL $\downarrow$
& 4.243 & 7.752
& 4.281 & 9.703
& 6.002 & 10.856 \\

\bottomrule
\vspace{-15pt}
\end{tabular}
\end{table}

\textbf{Cross-detector evasion.}
Table~\ref{tab:cross_detector} evaluates EASE against eight
additional detectors across three source LLMs.
Under fixed score orientations, EASE consistently reduces
AUROC and lowers or maintains TPR@1\%FPR across all
detector--model pairs.
The average TPR@1\%FPR across these eight detectors decreases
from 0.425 to 0.203 on Qwen3-8B, from 0.446 to 0.146 on
Llama3-8B, and from 0.141 to 0.046 on Ministral-3-8B.
The effect nevertheless varies across settings:
Likelihood and Binoculars show pronounced reductions in
low-FPR detection, whereas LRR retains substantial
discriminative ability.
Some detectors already perform near or below chance under
vanilla generation, so further AUROC reductions in these
settings should be interpreted under the fixed score orientation.
Overall, these results support the applicability of EASE across
different source LLMs and detectors.

\textbf{Generation quality.}
Table~\ref{tab:text_quality} evaluates text quality across three source LLMs.
BLEU-1 and ROUGE-1 remain relatively stable, while BLEU-2 and ROUGE-2
show small absolute decreases.
These results suggest limited degradation in reference overlap across
all three source LLMs, although PPL increases under EASE.

\subsection{Ablation Study and Inference Overhead}
\label{sec:ablation}

\textbf{Ablation Study.} 
Table~\ref{tab:ablation} examines the adaptive components of
EASE on Qwen3-8B.
Removing entropy adaptation fixes $\gamma_t=1$ at every step,
yielding constant logit perturbation strength and sampling
temperature, while removing temperature adaptation retains
entropy-guided logit shaping at the base temperature $\tau=1$.
Full EASE improves evasion over the fixed-strength variant
at similar PPL.
This result supports the benefit of using predictive entropy
to adapt logit perturbation strength and sampling temperature
across decoding steps.
Disabling temperature adaptation lowers PPL but weakens
evasion, showing that the temperature branch contributes
to attack effectiveness at an additional quality cost.
The complete configuration achieves the lowest detection
metrics among the evaluated variants, supporting the
effectiveness of the joint adaptive design.

\textbf{Inference overhead.}
On an RTX 3090, we benchmark Qwen3-8B in BF16 over 50 paired samples at batch size 1, 
generating 200 tokens from 30-token
prompts with $\tau=1$, $k=20$, and EASE $\delta=2$.
Excluding model loading, data I/O, and warm-up, EASE-Plugin
increases mean generation time from 12.362 to 12.442 s/sample,
an overhead of 0.643\%.
These results support its use as a plug-in decoding module
with little impact on generation latency.

\begin{table}[t]
\centering
\setlength{\tabcolsep}{4pt}
\renewcommand{\arraystretch}{1.05}
\caption{
Ablation of EASE on Qwen3-8B.
Detection metrics are averaged over the primary detectors shown in
Table~\ref{tab:main_competitors}.
}
\vspace{8pt}
\label{tab:ablation}

\begin{tabular}{lccc}
\toprule
\textbf{Variant}
& \textbf{AUROC}$\downarrow$
& \textbf{TPR@1\%FPR}$\downarrow$
& \textbf{PPL}$\downarrow$ \\
\midrule

Vanilla
& 0.952 & 0.655 & 4.243 \\

\midrule

w/o Entropy Adapt.
& 0.902 & 0.361 & 7.576 \\

w/o Temp. Adapt.
& 0.927 & 0.487 & \textbf{5.202} \\

\textbf{EASE}
& \textbf{0.879} & \textbf{0.339} & 7.752 \\

\bottomrule
\vspace{-15pt}
\end{tabular}
\end{table}

\section{Conclusion}
We proposed EASE, a training-free and detector-agnostic framework
that uses predictive entropy to adapt logit perturbation and sampling
temperature for evading AIGT detectors.
Our results show that changes in the source LLM's decoding distribution
can substantially reduce detection performance.
These findings highlight the need for detector designs that account
for decoding-induced changes in text statistics and remain robust
across different generation settings. 
Detector evaluation should therefore cover diverse decoding settings.


\section{COMPLIANCE WITH ETHICAL STANDARDS}
This is a computational study based on publicly available text data
and model-generated content. No human or animal subjects were involved,
and no ethical approval was required.

\bibliographystyle{IEEEbib}
\bibliography{mybib_final}

\end{document}